%% file: Lite-STGNN.tex
\documentclass[a4paper,twoside]{article}

\usepackage{epsfig}
\usepackage{subcaption}
\usepackage{calc}
\usepackage{amssymb}
\usepackage{amstext}
\usepackage{amsmath}
\usepackage{amsthm}
\usepackage{multicol}
\usepackage{pslatex}
\usepackage{apalike}
\usepackage{algorithm2e}
\usepackage[bottom]{footmisc}

\usepackage{booktabs}
\usepackage{threeparttable}
\usepackage[section]{placeins}
\usepackage{hyperref}
\usepackage{xcolor}

\usepackage{SCITEPRESS}
\usepackage{float}

\begin{document}

	\title{A lightweight Spatial-Temporal Graph Neural Network for Long-term Time Series Forecasting}

\author{\authorname{Henok Tenaw Moges\sup{1}\orcidAuthor{0000-0002-7158-8265}, and Deshendran Moodley\sup{1}\orcidAuthor{0000-0002-4340-9178}}
\affiliation{\sup{1}Centre for Artificial Intelligence Research (CAIR), \\ University of Cape Town, Cape Town, South Africa}
\email{\{henok.moges, deshen.moodley\}@uct.ac.za}
}


\abstract{We propose Lite-STGNN, a lightweight spatial-temporal graph neural network for long-term multivariate forecasting that integrates decomposition-based temporal modeling with learnable sparse graph structure. The temporal module applies trend-seasonal decomposition, while the spatial module performs message passing with low-rank Top-$K$ adjacency learning and conservative horizon-wise gating, enabling spatial corrections that enhance a strong linear baseline. Lite-STGNN achieves state-of-the-art accuracy on four benchmark datasets for horizons up to 720 steps, while being parameter-efficient and substantially faster to train than transformer-based methods. Ablation studies show that the spatial module yields 4.6\% improvement over the temporal baseline, Top-$K$ enhances locality by 3.3\%, and learned adjacency matrices reveal domain-specific interaction dynamics. Lite-STGNN thus offers a compact, interpretable, and efficient framework for long-term multivariate time series forecasting. The source code is available at \href{https://github.com/HTMoges/Lite-STGNN.git}{https://github.com/HTMoges/Lite-STGNN}.}

\maketitle \normalsize
\begingroup
    \renewcommand\thefootnote{}
    \footnotetext{\small \textit{Accepted at the 18th International Conference on Agents and Artificial Intelligence (ICAART 2026). Cite the proceedings/DOI when available.}}
\endgroup
\setcounter{footnote}{0} \vfill
\pagestyle{plain}

\section{\uppercase{Introduction}}
\label{sec:introduction}
Long-term multivariate time series forecasting (LTSF) is crucial for decision-making in domains such as energy grid management, traffic optimization, financial risk analysis, and climate prediction. Under long horizons (e.g., 336-720 steps), accurate forecasting requires modeling both intra-variable temporal dynamics and inter-variable dependencies. As the number of variables and forecast horizon increase, computational efficiency becomes critical. Linear models are stable and fast but often miss cross-variable structure, whereas transformer- and graph-based models capture interactions but at the cost of higher computational and parameter costs.

Spatial-temporal graph neural networks (STGNNs) such as DCRNN~\cite{li2017diffusion}, STGCN~\cite{yu2017spatio}, Graph WaveNet~\cite{wu2019graph}, AGCRN~\cite{bai2020adaptive}, and MTGNN~\cite{wu2020mtgnn} are strong at spatial modeling, but most are designed for short horizons (typically 3-24 steps) and become expensive or unstable as horizons grow. Lightweight variants~\cite{wang2025lightweight} improve efficiency, yet long-horizon validation remains limited.

In parallel, temporal modeling for LTSF has progressed along complementary lines. Linear decomposition-based models (DLinear~\cite{zeng2023dlinear}, RLinear~\cite{li2023rlinear}) provide strong long-horizon baselines, convolutional designs (ModernTCN~\cite{luo2024moderntcn}) improve temporal modeling through dilated convolutions, and transformer variants (PatchTST~\cite{nie2023patchtst}, TimesNet~\cite{wu2023timesnet}) extend receptive fields via patching or frequency representations. Despite these advances, many LTSF models process variables independently and underuse inter-variable structure.

\textbf{These observations motivate our primary question:} \emph{Can we embed learnable spatial dependencies within efficient temporal modules to achieve accurate, stable, and scalable long-term forecasting?}

To address this, we propose Lite-STGNN, a residual STGNN that augments a decomposition-based linear temporal backbone with a learnable sparse graph module. The spatial module uses low-rank adjacency factorization~\cite{wilson2018low} and Top-$K$ sparsification~\cite{plotz2024using}, reducing adjacency complexity from $O(N^2)$ to $O(Nr)$ ($r\ll N$). Conservative gating fuses spatial corrections as a residual refinement.

We evaluate Lite-STGNN under standard LTSF protocols~\cite{zeng2023dlinear,luo2024moderntcn,Zeng2024HowMC} on Electricity, Traffic, Exchange, and Weather. Lite-STGNN achieves state-of-the-art accuracy on Electricity and Exchange while remaining parameter-efficient (e.g., 0.74M parameters, 174$\times$ fewer than ModernTCN for Electricity) and training approximately 20$\times$ faster. The learned adjacency matrices are interpretable and align with domain structure.

In summary, Lite-STGNN bridges the gap between efficient linear forecasting and interpretable spatial-temporal graph learning, offering a practical, scalable, and clear solution for long-horizon multivariate forecasting.

\section{\uppercase{Related Work}} \label{sec:related_work}
Recent LTSF research largely splits into (i) temporal models that ignore spatial structure and (ii) STGNNs that model variable relationships explicitly but are typically validated on short horizons. We briefly review both directions and position Lite-STGNN at their intersection.

On the temporal side, early multivariate methods relied on recurrent architectures (LSTMs, GRUs), whose performance degrades over long sequences due to vanishing gradients. More recent work favors decomposition-based and convolutional approaches that explicitly separate trend and seasonal components. Representative examples include DLinear~\cite{zeng2023dlinear}, which decomposes series into independent linear projections and matches transformer performance with far fewer parameters; RLinear~\cite{li2023rlinear}, which shows that single-layer linear models with RevIN~\cite{kim2021reversible} scale to 720-step horizons; ModernTCN~\cite{luo2024moderntcn}, which uses dilated convolutions for long-range processing; and transformer variants such as PatchTST~\cite{nie2023patchtst}, TimesNet~\cite{wu2023timesnet}, and CycleNet~\cite{cheng2024cyclenet}, which use patching or frequency-domain representations. Despite their flexibility, transformers often incur high memory cost and millions of parameters, and all these models treat variables independently, neglecting inter-series dependencies that are crucial in domains such as energy systems and transportation networks.

On the spatial-temporal side, STGNNs model spatial dependencies using graph structure. Early works such as DCRNN~\cite{li2017diffusion} and STGCN~\cite{yu2017spatio} achieved success in short-term traffic prediction (3-12 steps) by coupling graph convolution with recurrent or convolutional temporal modules. Subsequent models, including Graph WaveNet~\cite{wu2019graph}, introduced self-adaptive adjacency learning, while AGCRN~\cite{bai2020adaptive} and AST-GCN~\cite{zhu2021ast} incorporated adaptive and attention-based spatial mechanisms. More recent developments such as DSTAGNN~\cite{lan2022dstagnn} and Dynamic Graph Neural ODEs~\cite{jin2022multivariate} represent temporal evolution as continuous dynamics. However, these architectures primarily target short-term horizons (12-24 steps) and often rely on heavy RNN or TCN temporal components, which limit scalability to the 96-720 step regime.

Recent studies have attempted to extend STGNNs towards greater scalability. BigST~\cite{han2024bigst} employs factorized graph representations to mitigate adjacency costs but remains primarily evaluated on traffic benchmarks with limited horizon length. Emerging lightweight STGNNs such as LSTNN~\cite{wang2025lightweight} demonstrate that decomposition-based or sampling-driven spatial designs can improve efficiency, yet these methods remain domain-specific and lack validation across diverse multivariate datasets. Low-rank factorization and graph pruning have also been explored to reduce computational complexity by retaining only the strongest edges. Low-rank adjacency factorization~\cite{wilson2018low} and Top-$K$ sparsification~\cite{plotz2024using} reduce adjacency complexity from $O(N^2)$ to $O(Nr)$, where $r \ll N$ denotes the embedding rank, enabling efficient message passing while preserving dominant spatial dependencies. In summary, while existing STGNNs effectively capture localized dependencies, they generally struggle to scale to extreme long-term horizons. Lite-STGNN fills this gap by unifying decomposition-based long-term temporal modeling with learnable sparse spatial reasoning, achieving scalability, interpretability, and competitive accuracy across 96-720 step LTSF.

\section{\uppercase{Proposed Approach}}\label{sec:method}
For long-term forecasting, scalability, interpretability, and stability are central challenges. Lite-STGNN addresses these through a lightweight spatial module that performs message passing using low-rank adjacency factorization \cite{wilson2018low} and Top-$K$ sparsification \cite{wu2019graph,plotz2024using} for efficient, interpretable graph learning. This reduces adjacency complexity from $O(N^2)$ to $O(Nr)$ and allows the spatial module to act as a residual correction to a decomposition-based temporal backbone, yielding measurable gains even at extended horizons.

Given a multivariate series $\mathbf{X} \in \mathbb{R}^{T \times N}$, where $T$ is the input length and $N$ is the number of features (e.g., sensors, nodes, variables), the goal is to predict the next $L$ future steps $\mathbf{Y} \in \mathbb{R}^{L \times N}$. Formally, we learn a mapping $f_{\Theta}: \mathbb{R}^{T \times N} \rightarrow \mathbb{R}^{L \times N}$ such that $\mathbf{Y} = f_{\Theta}(\mathbf{X}; \Theta)$, where $\Theta$ denotes all learnable parameters. The challenge is to efficiently capture both long-range temporal dependencies and cross-variable spatial interactions for horizons up to $L = 720$. To model spatial dependencies, each variable is treated as a node within a latent interaction structure whose weighted connections represent dynamic influences among variables, enabling message passing that captures both local intra-variable and global inter-variable dynamics~\cite{gaibie2024predicting}.

\subsection{Architecture}
\begin{figure}[t]
\centering
 \includegraphics[width=0.42\textwidth]{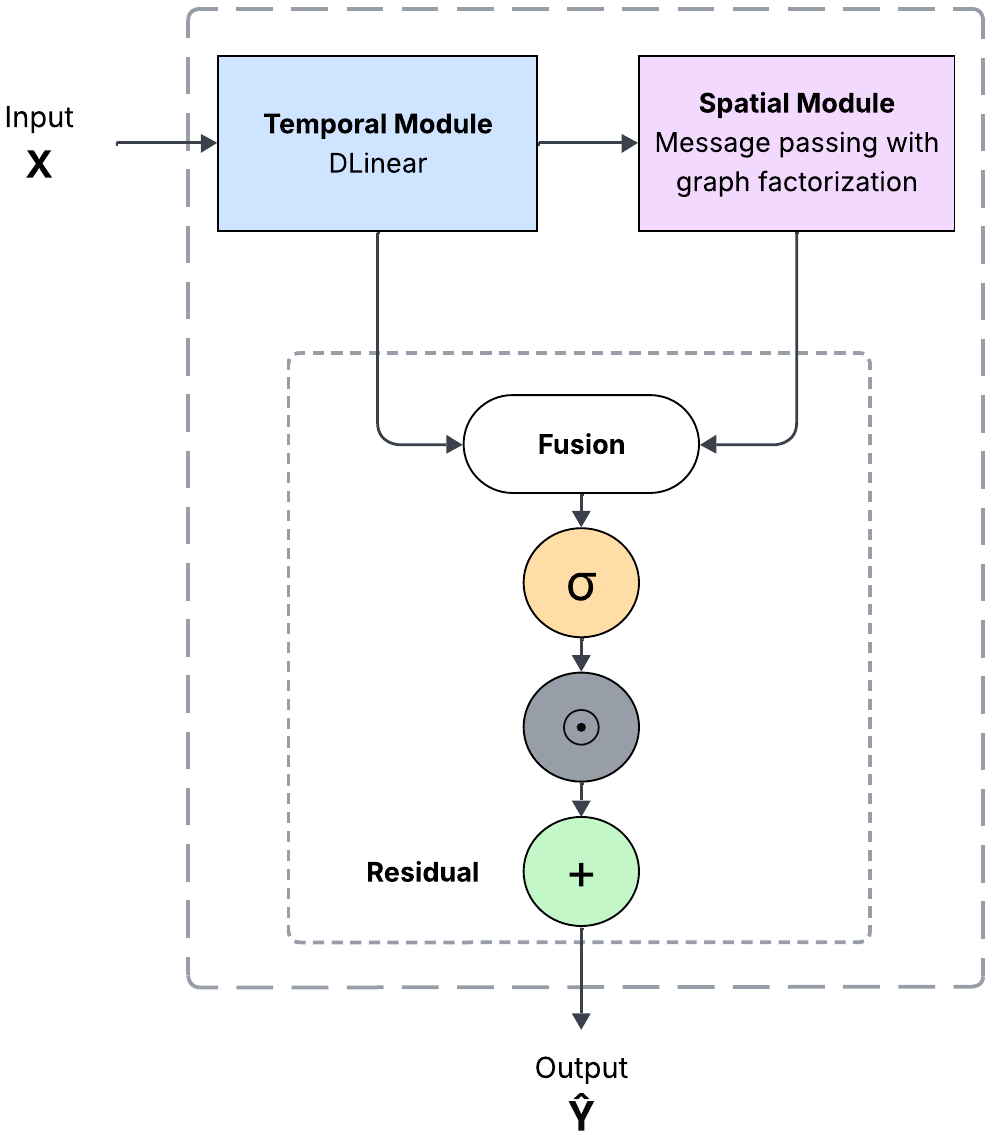}
 \captionsetup{font=small,skip=4pt}
\caption{Lite-STGNN architecture. A decomposition-based temporal module is combined with a lightweight \textbf{spatial module} that learns a low-rank Top-$K$ sparse adjacency to capture cross-variable dynamics. Conservative gating ($\sigma$), element-wise modulation ($\odot$), and residual addition ($+$) fuse spatial corrections with the temporal baseline.}
\label{fig:architecture}
\end{figure}

Figure~\ref{fig:architecture} presents the overall architecture of Lite-STGNN for efficient and interpretable long-horizon multivariate forecasting. A decomposition-based temporal module generates a strong base forecast $\mathbf{Y}_{\text{base}}$, a lightweight spatial module captures cross-variable dependencies through a low-rank Top-$K$ sparse adjacency, and residual fusion regulates spatial corrections via conservative gating ($\sigma$), element-wise modulation ($\odot$), and residual addition ($+$) to produce the final forecast $\mathbf{\hat{Y}}$. The final prediction is obtained as
\begin{equation}
\mathbf{\hat{Y}} = \mathbf{Y}_{\text{base}} + g(\Delta \mathbf{Y}).
\end{equation}
where $g(\cdot)$ denotes the gating function controlling the contribution of spatial dependencies.

\subsection{Temporal Module}
The temporal backbone follows the DLinear decomposition paradigm~\cite{zeng2023dlinear}. Each input sequence is decomposed into trend and seasonal components, $\mathbf{X} = \mathbf{X}^{\text{(trend)}} + \mathbf{X}^{\text{(season)}}$, where moving-average smoothing extracts the trend signal. Each component is then projected independently over the forecasting horizon, $\mathbf{Y}_{\text{base}} = \mathbf{W}_{\text{trend}}\mathbf{X}^{\text{(trend)}} + \mathbf{W}_{\text{season}}\mathbf{X}^{\text{(season)}}$, with learnable matrices $\mathbf{W}_{\text{trend}}, \mathbf{W}_{\text{season}} \in \mathbb{R}^{L \times T}$. This linear decomposition achieves $O(NL)$ complexity, offering strong performance and numerical stability.

\subsection{Spatial Module}
To capture spatial interactions among variables, Lite-STGNN performs message passing using a compact graph adjacency learned through low-rank factorization and Top-$K$ sparsification. Inspired by low-rank parametrizations used in \cite{wilson2018low} to reduce graph convolutional complexity, the adjacency matrix is defined as: 
\begin{equation}
\mathbf{A} = \mathrm{TopK}\big(\mathrm{ReLU}(\mathbf{E}_{\text{src}}\mathbf{E}_{\text{dst}}^{\top})\big),
\end{equation}
where $\mathbf{E}_{\text{src}}, \mathbf{E}_{\text{dst}} \in \mathbb{R}^{N \times r}$ are learnable node embeddings and $r \ll N$ is the embedding rank controlling the degree of compression. The low-rank construction provides a scalable approximation of pairwise node influence, reducing complexity from $O(N^2)$ to $O(Nr)$.  

Following the principle of adaptive graph structure introduced in Graph WaveNet \cite{wu2019graph}, the Top-$K$ operator retains only the strongest $k$ connections per node, filtering out weak or noisy correlations while preserving key structural relations \cite{plotz2024using}. The resulting adjacency matrix $\mathbf{A}$ is row-normalized to preserve numerical stability during aggregation.

The spatial residual is then computed as $\Delta \mathbf{Y} = (\mathbf{A} - \mathbf{I}) \mathbf{Y}_{\text{base}}$, where $\mathbf{I}$ removes self-influence, isolating cross-variable effects and allowing message passing to act as a dynamic correction mechanism.

\subsection{Residual Fusion and Gating}
Spatial corrections are fused as a conservative residual rather than replacing the temporal forecast. The gating mechanism is defined as:

\begin{equation}
g(\Delta \mathbf{Y}) = \beta\, \sigma(\mathbf{w}_{\text{gate}}) \odot \Delta \mathbf{Y},
\end{equation}
where $\sigma(\cdot)$ is the sigmoid, $\mathbf{w}_{\text{gate}}\!\in\!\mathbb{R}^{L}$ are per-horizon biases, and $\beta=\text{softplus}(\theta_{\text{scale}})$ ensures positive scaling. We initialize $\mathbf{w}_{\text{gate}}=-4.0$, giving $\sigma(-4.0)\approx0.018$, so spatial corrections initially contribute only $\sim\!2\%$ of their magnitude and the gates learn when spatial signals are useful.

This residual design preserves stability by allowing the model to stay close to the temporal baseline when spatial dependencies are weak.

\subsection{Training Strategy}
Lite-STGNN is trained using mean squared error (MSE) as the objective, following the common experimental pipeline used in prior long-term forecasting benchmarks such as DLinear~\cite{zeng2023dlinear} and ModernTCN~\cite{luo2024moderntcn}. We report both MSE and MAE, use early stopping with patience=10 based on mean validation MSE across horizons (96, 192, 336, 720), and evaluate each configuration over three independent runs. All experiments are implemented in PyTorch, and trained on a single NVIDIA RTX~3090 GPU.

\section{\uppercase{Results and Discussion}}\label{sec:results}
This section evaluates Lite-STGNN against six baselines: DLinear~\cite{zeng2023dlinear}, RLinear~\cite{li2023rlinear}, PatchTST~\cite{nie2023patchtst}, TimesNet~\cite{wu2023timesnet}, CycleNet~\cite{cheng2024cyclenet}, and ModernTCN~\cite{luo2024moderntcn} on four benchmarks: Electricity, Exchange, Traffic, and Weather. All models share the same training protocol, and we report both MSE and MAE.

\subsection{Quantitative Comparison}
\begin{figure*}[!t]
\centering
\includegraphics[width=0.6\textwidth]{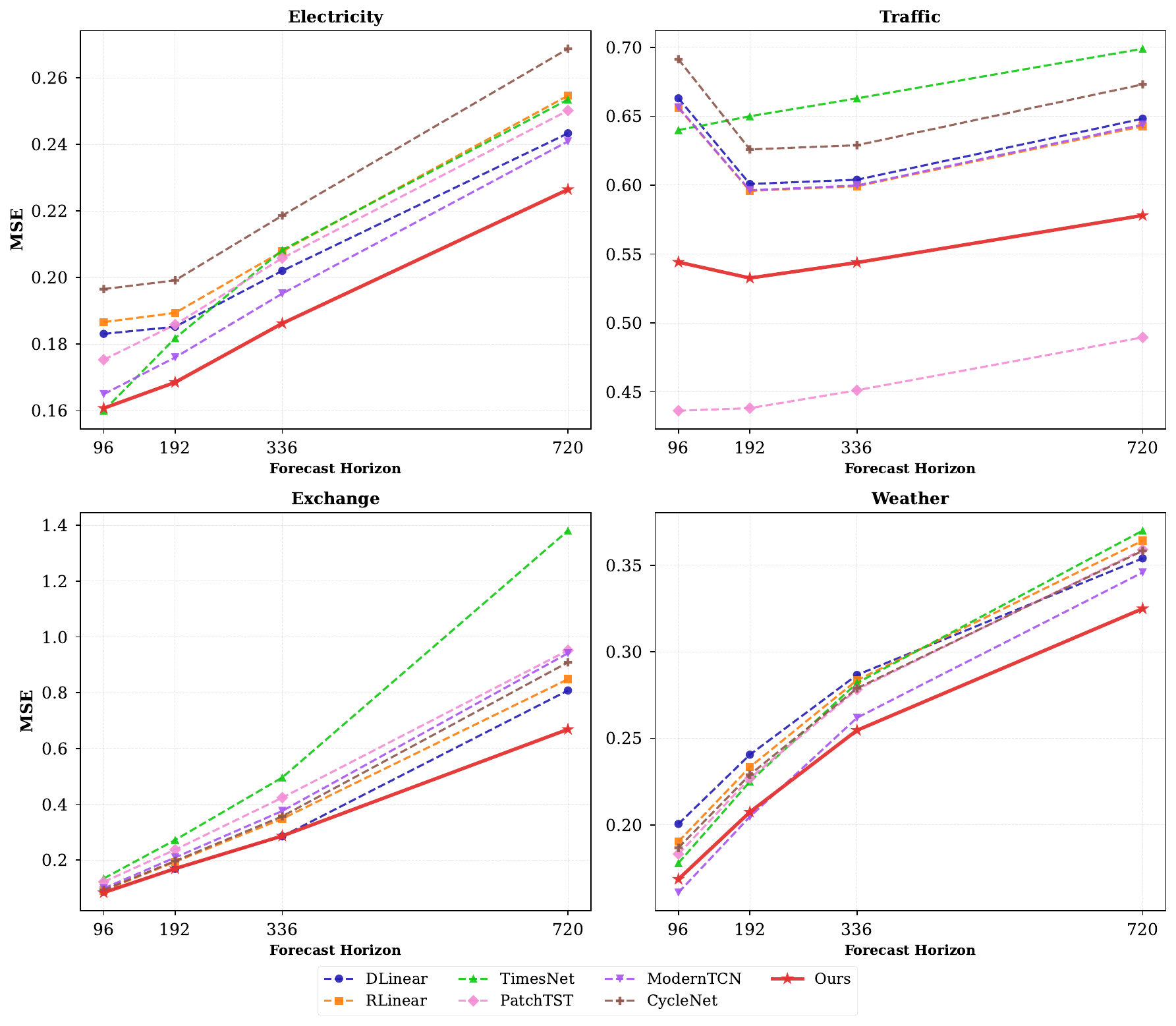}
\captionsetup{font=small,skip=4pt}
\caption{Per-horizon MSE over horizons 96/192/336/720 (mean over 3 runs). Lite-STGNN remains among the top performers, especially on Electricity and Exchange (Table~\ref{tab:main_results}).}
\label{fig:per_horizon}
\end{figure*}

Table~\ref{tab:main_results} presents quantitative results across four forecasting horizons (96, 192, 336, and 720). Lite-STGNN consistently achieves state-of-the-art performance compared to baselines in both MSE and MAE metrics. Figure~\ref{fig:per_horizon} visualizes the corresponding error trends across horizons.

\input{tables/main_results_per_horizon}

Across all datasets and horizons, Lite-STGNN attains the best or second-best performance. Error growth with horizon is noticeably smaller than for most baselines, indicating that residual spatial correction improves long-term stability without compromising computational efficiency. Key findings per dataset are:  

\begin{itemize}
\item \textit{Electricity}: Lite-STGNN achieves best performance in both MSE and MAE, with mean MSE of 0.178 (vs ModernTCN~0.194, +8.2\%) and MAE of 0.280 (vs ModernTCN~0.284, +1.4\%), indicating that the model maintains temporal structure while exploiting inter-sensor dependencies.
\item \textit{Exchange}: Lite-STGNN achieves the best long-horizon performance overall, with mean MSE=0.362 and MAE=0.369. While DLinear is slightly better at the shortest horizons (H96/H192), Lite-STGNN clearly dominates at longer ranges. At H720, for example, Lite-STGNN attains MSE=0.552 and MAE=0.583, outperforming DLinear's 0.722 and 0.667 (+23.5\% MSE and +12.6\% MAE). Despite weaker direct spatial coupling among currencies, the learned adjacency captures latent dependencies, yielding a smoother MSE curve in Fig.~\ref{fig:per_horizon}.
\item \textit{Traffic}: On this highly dynamic dataset ($N=862$), PatchTST attains the best accuracy (MSE 0.454, MAE 0.295), while Lite-STGNN achieves the second-best MSE=0.552 and MAE=0.328, outperforming ModernTCN by 11.6\% in MAE (0.328 vs 0.371) and surpassing the remaining baselines including DLinear, RLinear, TimesNet, and CycleNet.
\item \textit{Weather}: Lite-STGNN maintains a consistent advantage across all horizons, obtaining the best MSE (0.239) among all baselines and outperforming ModernTCN (0.243). Its MAE=0.294 is competitive but slightly higher than ModernTCN's 0.278, while remaining comparable to other strong baselines.
\end{itemize}

Overall, Lite-STGNN achieves state-of-the-art results on Electricity and Exchange in both MSE and MAE, and ranks among the top two models across all datasets while remaining parameter-efficient. Figure~\ref{fig:per_horizon} shows per-horizon MSE for horizons 96, 192, 336, and 720. On Exchange, the most volatile dataset, Lite-STGNN improves MSE by +23.5\% over DLinear at H720, highlighting the benefit of learned spatial message passing for long-range dependencies and stable error growth at extended horizons.

\subsection{Parameter and Computational Efficiency}
A key advantage of Lite-STGNN is its parameter efficiency combined with fast training. On the Electricity dataset (Table~\ref{tab:param_efficiency_electricity}), the learned spatial module adds only $\approx$0.69M parameters to a baseline temporal module (0.14M), yet achieves state-of-the-art performance with MSE=0.178 and MAE=0.280. This outperforms ModernTCN (129.3M parameters, MAE 0.284, MSE 0.194) by 8.2\% MSE and 1.4\% MAE, while using 174$\times$ fewer parameters and training in 27.3 s/epoch vs $\approx$545.2 s/epoch.

\input{tables/param_efficiency_with_time}

Lite-STGNN scales linearly with the number of nodes and horizon length, with overall complexity $O(Nr + NL)$. For example, on the Electricity dataset ($N=321$, $T=96$, $L=720$), the total parameter count is only $\approx$0.74M, compared to 129M and 7.6M for ModernTCN and PatchTST, respectively. The low-rank Top-$K$ design therefore achieves an effective balance between accuracy and scalability across both small- and large-scale benchmarks.

\subsection{Learned Spatial Structure and Forecasting Visualizations}

Figure~\ref{fig:adjacency_all} visualizes the learned adjacency matrices for all four datasets at horizon H720, with each cell $(i, j)$ denoting the learned edge weight between nodes $i$ and $j$. The learned adjacency patterns reveal interpretable domain structures:
\begin{itemize}
\item \textbf{Electricity}: The sparse network exhibits distinct regional clusters corresponding to geographic power grid zones, with strong edges forming coherent communities rather than diffuse patterns.
\item \textbf{Exchange}: Despite the small $N=8$, the model identifies strong links between currency pairs with known macroeconomic coupling (e.g., EUR-USD, GBP-USD), while peripheral currencies show weaker dependencies.
\item \textbf{Traffic}: The adjacency reveals corridor-like structures reflecting physical road networks, where sensors along the same corridor show strong upstream-downstream influence.
\item \textbf{Weather}: Weather variables show widespread dynamic coupling driven by atmospheric transport, with the strongest interactions following spatial proximity.
\end{itemize}

These results indicate that Lite-STGNN uncovers dynamic relationships that drive system evolution at long horizons, supporting interpretability and validating the model's domain-aware behaviour. Furthermore, Figure~\ref{fig:adjacency_all} shows a representative run; hence, similar structures can be observed across independent runs.

\begin{figure}[!t]
\centering
\begin{subfigure}[b]{\linewidth}
    \centering
    \includegraphics[width=0.95\textwidth]{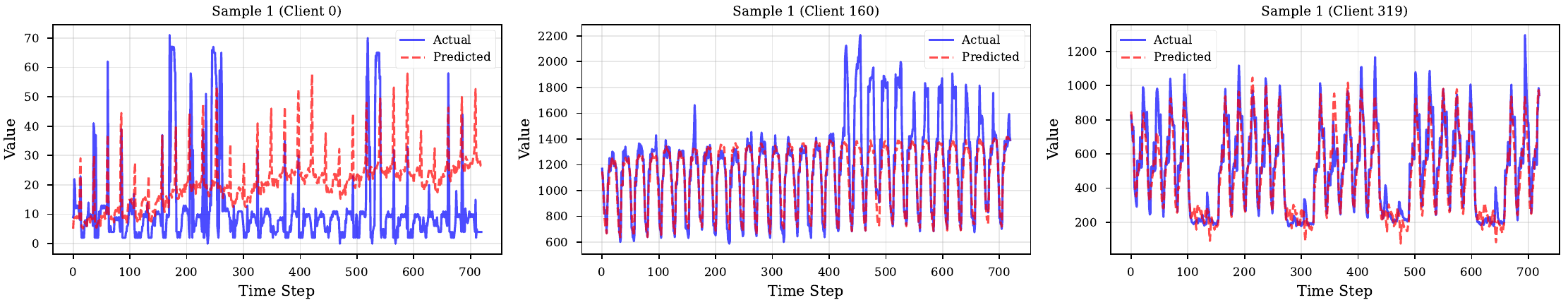}
    \caption{Electricity predictions at H720}
    \label{fig:predictions_electricity}
\end{subfigure}
\\[2mm]
\begin{subfigure}[b]{\linewidth}
    \centering
    \includegraphics[width=0.95\textwidth]{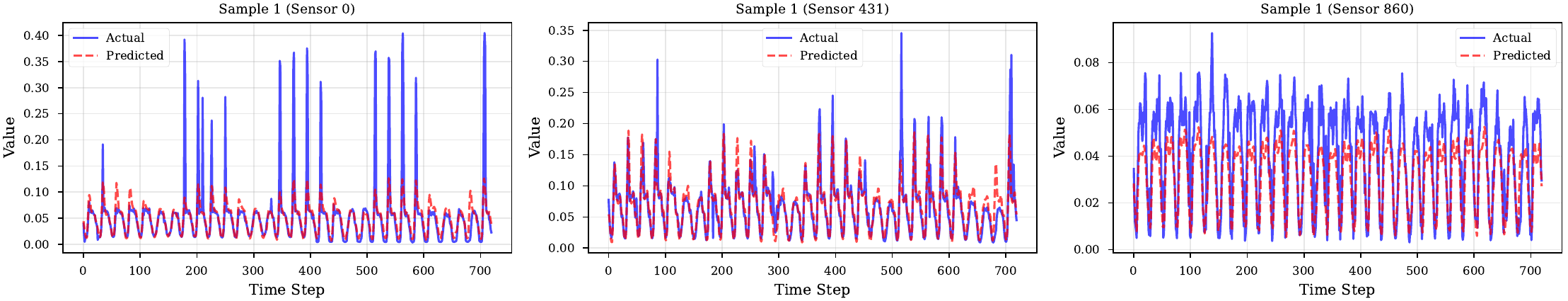}
    \caption[d]{Traffic predictions at H720}
    \label{fig:predictions_traffic}
\end{subfigure}
\captionsetup{font=small,skip=4pt}
\caption{Representative forecast samples at H720. We observe similar results for other samples, confirming that these cases are representative.}
\label{fig:predictions_all}
\end{figure}

Figure~\ref{fig:predictions_all} presents representative forecast samples at the 720-step horizon on Electricity and Traffic, where the close alignment between predicted and actual trajectories illustrates that Lite-STGNN captures both short-term transitions and long-range dependencies. We observe similar results for other samples and datasets (Exchange, Weather), confirming that these visualizations are representative of the model's general behavior.

\begin{figure}[ht]
\centering
\captionsetup{font=small,skip=4pt}
\begin{subfigure}[!hb]{0.45\textwidth}
    \includegraphics[width=\textwidth]{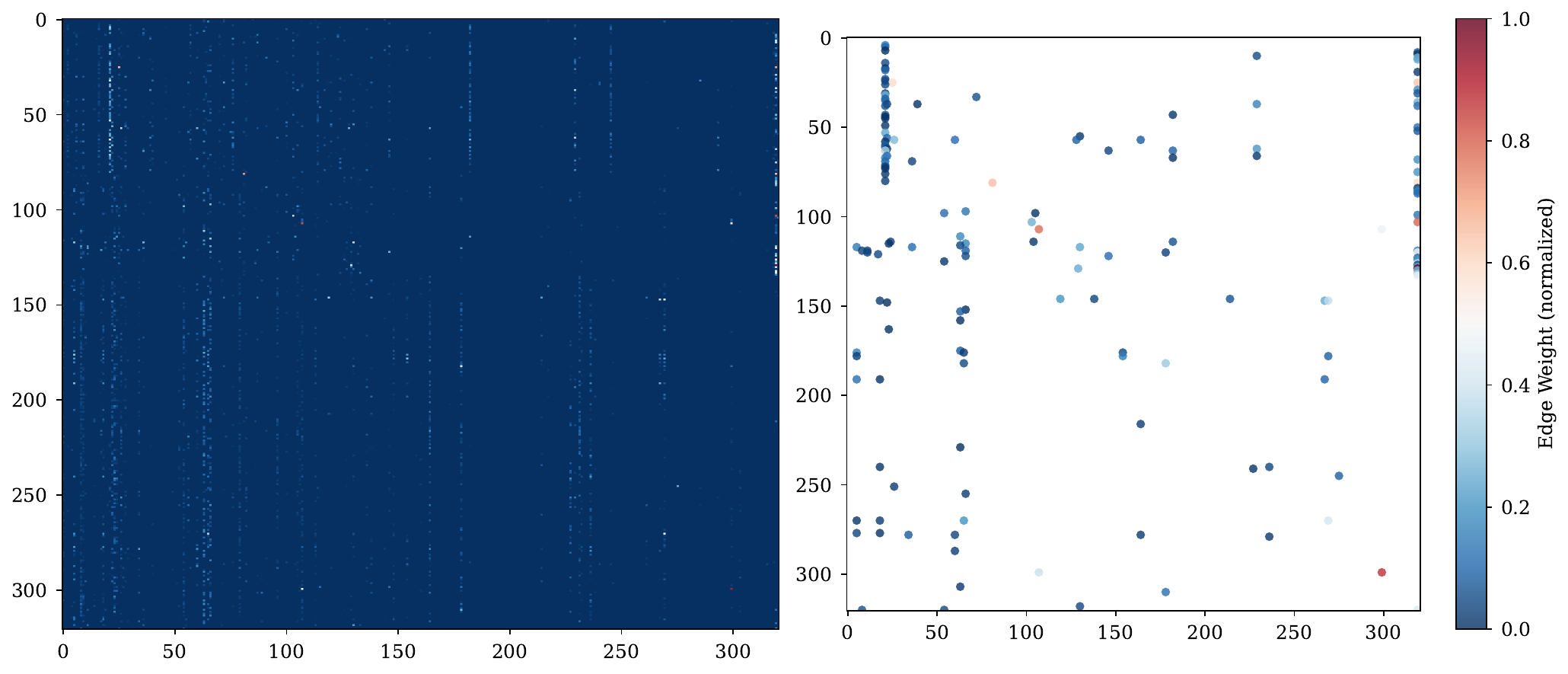}
    \caption{Electricity ($N=321$)}
    \label{fig:adjacency_electricity}
\end{subfigure}
\vspace{4pt}

\begin{subfigure}[!hb]{0.45\textwidth}
    \includegraphics[width=\textwidth]{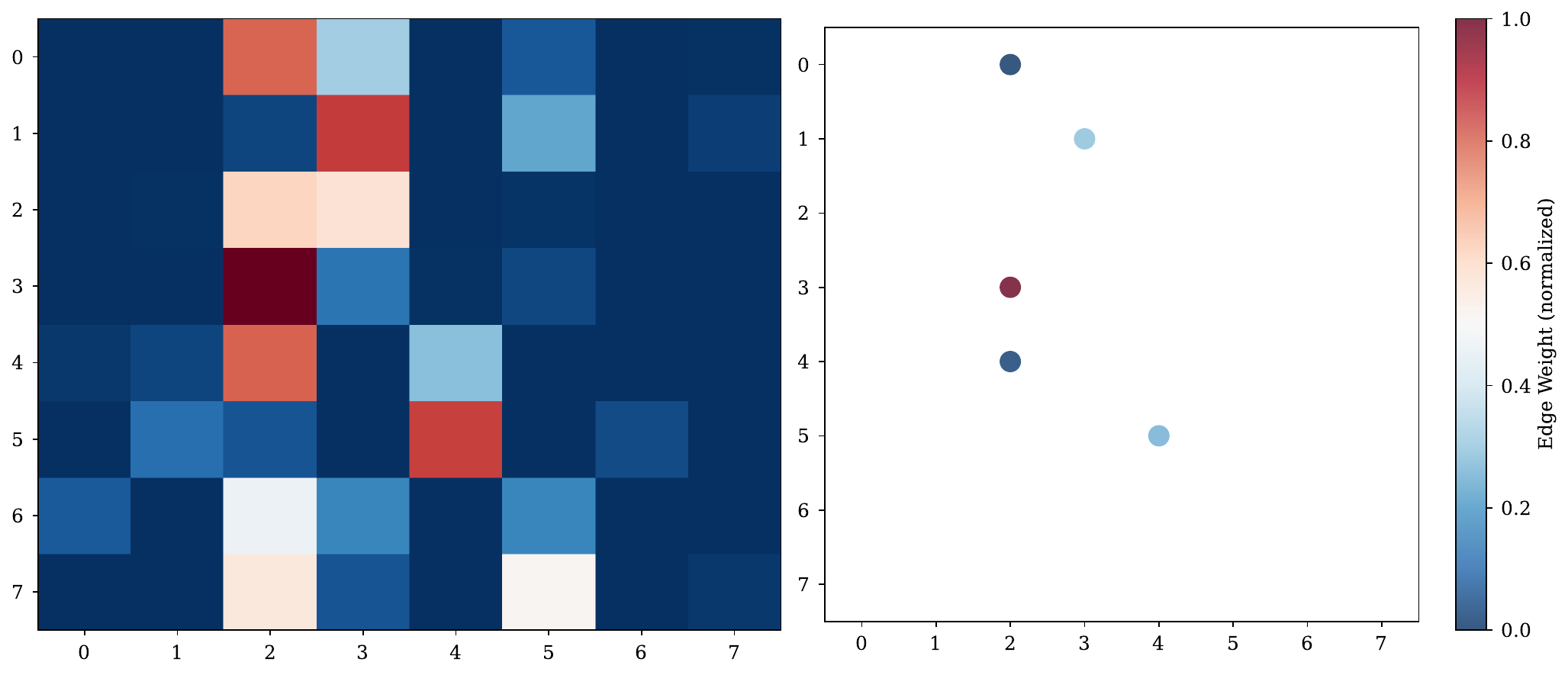}
    \caption{Exchange ($N=8$)}
    \label{fig:adjacency_exchange}
\end{subfigure}
\vspace{4pt}

\begin{subfigure}[!hb]{0.45\textwidth}
    \includegraphics[width=\textwidth]{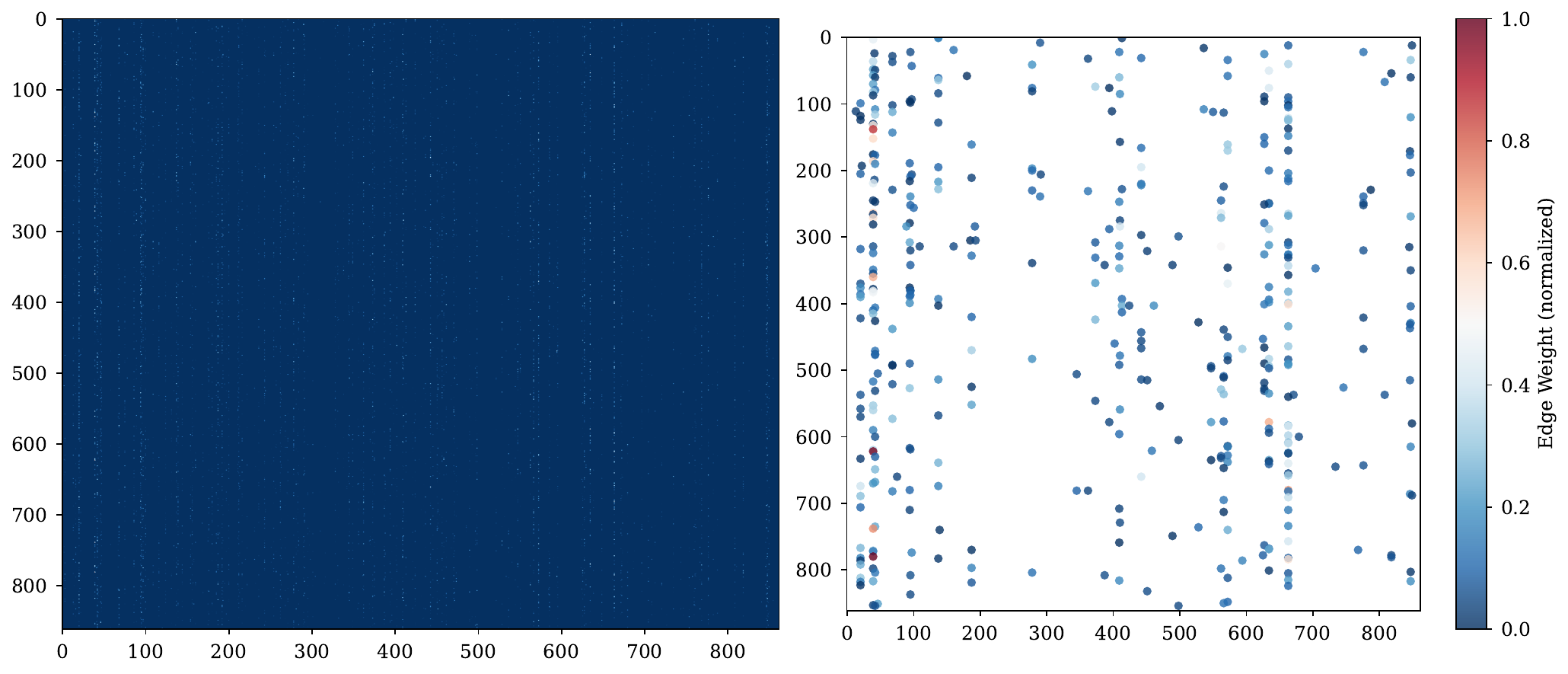}
    \caption{Traffic ($N=862$)}
    \label{fig:adjacency_traffic}
\end{subfigure}
\phantomcaption
\end{figure}

\begin{figure}[t]
\ContinuedFloat
\centering
\captionsetup{font=small,skip=4pt}
\begin{subfigure}[!hb]{0.45\textwidth}
    \includegraphics[width=\textwidth]{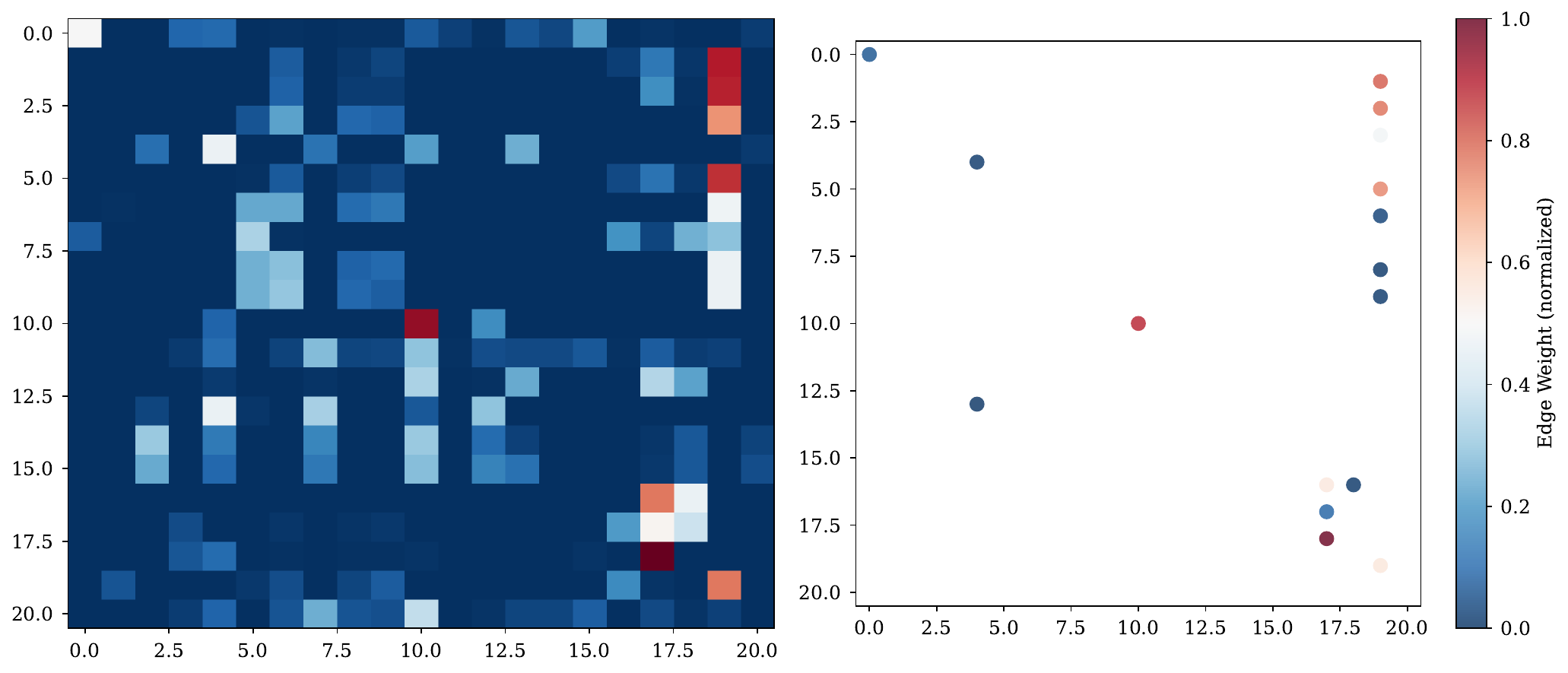}
    \caption{Weather ($N=21$)}
    \label{fig:adjacency_weather}
\end{subfigure}
\caption{Learned adjacency matrices (normalized weights) at horizon H720. Left: full connectivity; right: top 5\% strongest edges.}
\label{fig:adjacency_all}
\end{figure}

\section{\uppercase{Ablation Studies}}\label{sec:ablations}
We conduct comprehensive ablation studies on the Electricity dataset as an illustrative case to validate the proposed architectural design and to quantify the contributions of each model component.

\subsection{Spatial Module Contribution}
To assess the impact of modeling cross-variable spatial dependencies, we evaluate ten systematically selected architectural configurations on the Electricity dataset with multi-seed validation (3 independent runs). The variations include: temporal-only (DLinear), adjacency parameterization (rank, Top-$K$), temperature parameter ($\tau$), propagation depth (prop), gating mode (flat vs. band), and regularization (Dropout levels). Figure~\ref{fig:ablations_electricity} presents the per-horizon MSE and MAE trajectories for all configurations across the four horizons.

Introducing the spatial module results in a 4.6\% improvement (MAE: 0.2939 to 0.280±0.007) over the DLinear-only baseline, validating the benefit of incorporating spatial interactions. Furthermore, the inclusion of Top-$K$ sparsification provides additional gains beyond the low-rank model alone (mean MSE=0.185±0.009 for rank=16, Top-$K$=10), demonstrating the effectiveness of our two-stage design: the low-rank formulation reduces complexity from $O(N^2)$ to $O(Nr)$, while Top-$K$ selection enforces sparse, localized connectivity. This combination stabilizes training and preserves meaningful inter-variable dynamics over longer horizons.

These results confirm that the Lite-STGNN design, comprising a low-rank adjacency formulation with adaptive Top-$K$ sparsification, achieves a favorable trade-off between model expressivity and computational efficiency.

\subsection{Adjacency Parameterization}
We evaluate sensitivity to rank $r$, sparsity $k$, temperature $\tau$, and propagation depth (prop) on Electricity. Multi-seed results indicate that \textbf{rank=16} yields the best overall accuracy (mean MSE=0.185±0.009; MAE=0.284±0.007), while rank=8 and rank=32 are slightly worse, suggesting that moderate embedding capacity is sufficient. For sparsity, \textbf{$k=10$} performs best (about 3\% density for $N=321$), whereas denser graphs (e.g., $k=15$) introduce noise and degrade performance. The model is relatively robust to $\tau\in\{0.8,1.0,1.2\}$, and \textbf{2-hop propagation} is necessary: prop=1 degrades accuracy, while prop=3 shows diminishing returns. The per-horizon trends are shown in Figure~\ref{fig:ablations_electricity}.

\subsection{Gating and Regularization}
\begin{figure*}[t]
\centering
\includegraphics[width=0.7\textwidth]{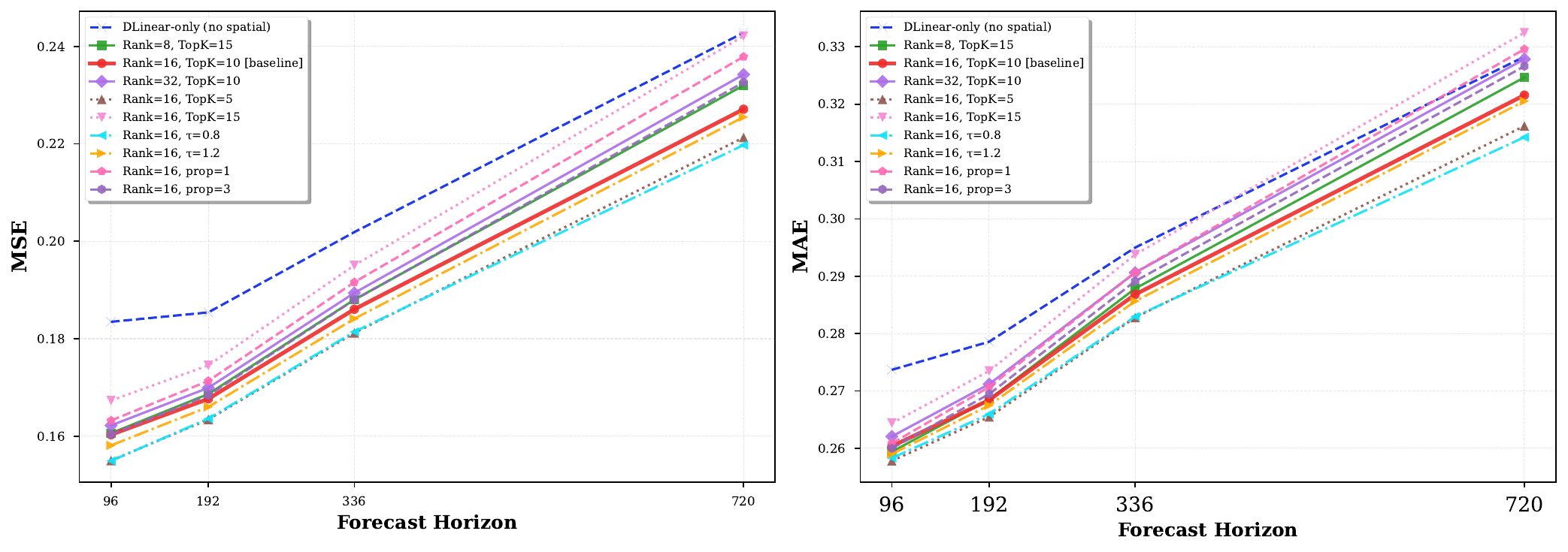}
\captionsetup{font=small,skip=4pt}
\caption{Ablation on Electricity (mean over 3 runs): spatial module improves MAE by 4.6\% over temporal-only; rank=16 and Top-$K$=10 perform best and remain stable across horizons.}
\label{fig:ablations_electricity}
\end{figure*}

We further examine two architectural factors: (1) gating mode (\textit{band} vs \textit{flat}) and (2) residual Dropout level. The results are summarized in Figure~\ref{fig:ablations_electricity}, which also visualizes the corresponding per-horizon MSE and MAE trends from the previous ablation study.

\section{\uppercase{Conclusions and Future Work}}\label{sec:conclusion}
This work introduced the \textbf{Lite-STGNN}, a lightweight spatial-temporal graph framework that balances accuracy, efficiency, and interpretability for long-term multivariate forecasting (96-720 steps). By integrating a strong decomposition-based linear model with a low-rank adaptive spatial module, Lite-STGNN achieves state-of-the-art performance across diverse datasets.

On Electricity and Exchange benchmarks, Lite-STGNN attains the best MSE and MAE (MSE=0.178, MAE=0.280), outperforming ModernTCN by +8.2\% MSE while using 174$\times$ fewer parameters and training 20$\times$ faster per epoch. Across all datasets, the model consistently ranks within the top two performers, demonstrating robust generalization from energy systems to financial markets.

Ablation studies validate the core design choices: the spatial module provides a 4.6\% improvement over the temporal baseline (DLinear), while Top-$K$ sparsity adds a further 3.3\% gain by concentrating information flow on the most relevant local interactions. The resulting adjacency patterns align with domain knowledge, such as regional power zones, currency coupling, and geographic sensor clusters, and the low-rank Top-$K$ spatial formulation scales as $O(Nr)$.

Overall, these findings confirm that \textit{adaptive spatial structure learning} offers a more efficient and principled route to improving long-term forecasting than brute-force parameter scaling, yielding a practical and resource-efficient framework for real-world spatial-temporal forecasting tasks.

Future work includes \emph{probabilistic forecasting} with calibrated uncertainty, \emph{dynamic adjacency} to capture regime shifts, and \emph{larger-scale validation} on explicitly spatial benchmarks.

\section*{\uppercase{Acknowledgements}}
H.~T.~M. and D.~M. acknowledge the financial support by the National Research Foundation (NRF) of South Africa (Grant No. 151217) and the use of the University of Cape Town High Performance Computing (HPC) facility for the experiments.

\small
\bibliographystyle{apalike}
\bibliography{main}

\end{document}

%% file: tables/main_results_per_horizon.tex
\begin{table*}[t!]
\centering
\captionsetup{font=small,skip=4pt}
\caption{Main results on four benchmarks. We report MAE and MSE at horizons 96, 192, 336, and 720, plus the mean across horizons. Best results are shown in \textbf{bold}, while the second best are \underline{underlined}.}
\label{tab:main_results}
\vspace{-1mm}
\resizebox{0.97\textwidth}{!}{
\renewcommand{\arraystretch}{0.85}
\begin{tabular}{l|cc|cc|cc|cc|cc}
\toprule
\textbf{Models} & \multicolumn{2}{c|}{\textbf{96}} & \multicolumn{2}{c|}{\textbf{192}} &
\multicolumn{2}{c|}{\textbf{336}} & \multicolumn{2}{c|}{\textbf{720}} &
\multicolumn{2}{c}{\textbf{Mean}} \\
& \textbf{MAE} & \textbf{MSE} & \textbf{MAE} & \textbf{MSE} &
\textbf{MAE} & \textbf{MSE} & \textbf{MAE} & \textbf{MSE} &
\textbf{MAE} & \textbf{MSE} \\
\midrule
\multicolumn{11}{c}{\textbf{Electricity}} \\
\midrule
CycleNet & 0.288 & 0.197 & 0.293 & 0.199 & 0.310 & 0.219 & 0.347 & 0.269 & 0.309 & 0.221 \\
DLinear & 0.272 & 0.183 & 0.278 & 0.185 & 0.295 & 0.202 & 0.329 & 0.243 & 0.294 & 0.203 \\
ModernTCN & \underline{0.264} & \underline{0.165} & \underline{0.272} & \underline{0.176} & \underline{0.289} & \underline{0.195} & 0.322 & 0.241 & \underline{0.287} & 0.195 \\
PatchTST & 0.278 & 0.168 & 0.287 & 0.178 & 0.304 & 0.199 & 0.335 & 0.242 & 0.301 & 0.197 \\
RLinear & 0.271 & 0.187 & 0.275 & 0.189 & 0.292 & 0.208 & 0.326 & 0.255 & 0.291 & 0.210 \\
TimesNet & 0.274 & 0.165 & 0.284 & 0.179 & 0.297 & 0.196 & \underline{0.320} & \underline{0.235} & 0.294 & \underline{0.194} \\
\textbf{Ours} & \textbf{0.259} & \textbf{0.156} & \textbf{0.267} & \textbf{0.164} & \textbf{0.282} & \textbf{0.180} & \textbf{0.311} & \textbf{0.214} & \textbf{0.280} & \textbf{0.178} \\
\midrule
\multicolumn{11}{c}{\textbf{Exchange}} \\
\midrule
CycleNet & 0.206 & 0.093 & 0.307 & 0.197 & 0.426 & 0.357 & 0.717 & 0.909 & 0.414 & 0.389 \\
DLinear & \textbf{0.199} & \textbf{0.086} & \textbf{0.288} & \textbf{0.168} & \underline{0.393} & \underline{0.285} & \underline{0.667} & \underline{0.808} & \underline{0.387} & \underline{0.337} \\
ModernTCN & 0.216 & 0.098 & 0.318 & 0.209 & 0.438 & 0.376 & 0.732 & 0.942 & 0.426 & 0.406 \\
PatchTST & 0.221 & 0.105 & 0.331 & 0.228 & 0.473 & 0.435 & 0.762 & 0.999 & 0.446 & 0.442 \\
RLinear & \underline{0.204} & 0.091 & 0.304 & 0.194 & 0.420 & 0.348 & 0.694 & 0.848 & 0.406 & 0.370 \\
TimesNet & 0.260 & 0.134 & 0.367 & 0.272 & 0.496 & 0.496 & 0.838 & 1.381 & 0.490 & 0.571 \\
\textbf{Ours} & 0.210 & \underline{0.088} & \underline{0.296} & \underline{0.170} & \textbf{0.386} & \textbf{0.278} & \textbf{0.583} & \textbf{0.594} & \textbf{0.369} & \textbf{0.282} \\
\midrule
\multicolumn{11}{c}{\textbf{Traffic}} \\
\midrule
CycleNet & 0.423 & 0.691 & 0.392 & 0.626 & 0.395 & 0.629 & 0.419 & 0.673 & 0.407 & 0.655 \\
DLinear & 0.399 & 0.663 & 0.371 & 0.601 & 0.374 & 0.604 & 0.399 & 0.648 & 0.386 & 0.629 \\
ModernTCN & 0.391 & 0.661 & 0.364 & 0.601 & 0.366 & 0.604 & 0.388 & 0.647 & 0.377 & 0.628 \\
PatchTST & \textbf{0.298} & \textbf{0.446} & \textbf{0.298} & \textbf{0.447} & \textbf{0.305} & \textbf{0.460} & \textbf{0.326} & \textbf{0.498} & \textbf{0.307} & \textbf{0.463} \\
RLinear & 0.384 & 0.656 & 0.356 & 0.596 & 0.359 & 0.599 & 0.382 & 0.643 & 0.370 & 0.623 \\
TimesNet & 0.329 & 0.640 & 0.332 & 0.650 & 0.339 & 0.663 & 0.359 & 0.699 & 0.340 & 0.663 \\
\textbf{Ours} & \underline{0.325} & \underline{0.545} & \underline{0.319} & \underline{0.536} & \underline{0.325} & \underline{0.547} & \underline{0.344} & \underline{0.580} & \underline{0.328} & \underline{0.552} \\
\midrule
\multicolumn{11}{c}{\textbf{Weather}} \\
\midrule
CycleNet & 0.231 & 0.187 & 0.267 & 0.229 & 0.302 & 0.279 & 0.349 & 0.358 & 0.287 & 0.263 \\
DLinear & 0.272 & 0.201 & 0.310 & 0.241 & 0.347 & 0.287 & 0.395 & 0.354 & 0.331 & 0.270 \\
ModernTCN & \textbf{0.214} & \textbf{0.161} & \textbf{0.255} & \textbf{0.205} & \textbf{0.296} & \underline{0.262} & \underline{0.348} & \underline{0.346} & \textbf{0.278} & \underline{0.243} \\
PatchTST & \underline{0.225} & 0.178 & \underline{0.262} & 0.222 & \underline{0.298} & 0.275 & \textbf{0.348} & 0.356 & \underline{0.283} & 0.258 \\
RLinear & 0.234 & 0.190 & 0.270 & 0.233 & 0.305 & 0.284 & 0.353 & 0.364 & 0.291 & 0.268 \\
TimesNet & 0.232 & 0.178 & 0.272 & 0.225 & 0.310 & 0.282 & 0.361 & 0.370 & 0.294 & 0.264 \\
\textbf{Ours} & 0.237 & \underline{0.171} & 0.273 & \underline{0.209} & 0.309 & \textbf{0.256} & 0.356 & \textbf{0.322} & 0.294 & \textbf{0.239} \\
\bottomrule
\end{tabular}
}
\vspace{-2mm}
\end{table*}

%% file: tables/param_efficiency_with_time.tex
\begin{table}[ht]
\centering
\captionsetup{font=small,skip=4pt}
\caption{Parameter and computational efficiency on the Electricity dataset. Lite-STGNN achieves the best MAE and MSE with competitive training time and $\mathbf{174\times}$ fewer parameters than ModernTCN.}
\label{tab:param_efficiency_electricity}
\vspace{-1mm}
\resizebox{\columnwidth}{!}{
\renewcommand{\arraystretch}{0.85}
\begin{tabular}{lcccc}
\toprule
\textbf{Model} & \textbf{Parameters} & \textbf{Train Time/Epoch (s)} & \textbf{Mean MAE} & \textbf{Mean MSE} \\
\midrule
CycleNet & 78K & 45.2 & 0.309 & 0.221 \\
DLinear & 0.14M & 15.7 & 0.294 & 0.203 \\
RLinear & 70K & 18.7 & 0.291 & 0.210 \\
\midrule
ModernTCN & 129.3M & 545.2 & 0.284 & 0.194 \\
PatchTST & 7.6M & 284.3 & 0.310 & 0.204 \\
TimesNet & 150.4M & 3510.6 & 0.294 & 0.194 \\
\midrule
\textbf{Ours} & \textbf{0.74M} & \textbf{27.3} & \textbf{0.280} & \textbf{0.178} \\
\bottomrule
\end{tabular}}
\vspace{-2mm}
\end{table}